\documentclass[10pt, a4paper]{article}
\usepackage{lrec2022} 
\usepackage{multibib}
\newcites{languageresource}{Language Resources}
\usepackage{graphicx}
\usepackage{tabularx}
\usepackage{soul}
\usepackage{titlesec}
\usepackage{amssymb}
\usepackage{amsmath}
\usepackage{multicol}
\usepackage{enumitem} 
\usepackage[utf8]{inputenc} 
\usepackage[LFE,LAE,T1]{fontenc} 
\usepackage{arabtex} 
\usepackage{utf8} 
\setcode{utf8} 

\usepackage{expex} 
\lingset{everygla=, belowglpreambleskip=-0.5ex, aboveglftskip=-0.5ex} 

\titleformat{\section}{\normalfont\large\bfseries\center}{\thesection.}{1em}{}
\titleformat{\subsection}{\normalfont\SmallTitleFont\bfseries\raggedright}{\thesubsection.}{1em}{}
\titleformat{\subsubsection}{\normalfont\normalsize\bfseries\raggedright}{\thesubsubsection.}{1em}{}
\renewcommand\thesection{\arabic{section}}
\renewcommand\thesubsection{\thesection.\arabic{subsection}}
\renewcommand\thesubsubsection{\thesubsection.\arabic{subsubsection}}

\usepackage{epstopdf}
\usepackage{xstring}

\usepackage{color}

\usepackage{tikz}
\usetikzlibrary{matrix,chains,positioning,decorations.pathreplacing,arrows}
\usetikzlibrary{positioning,calc}
\usetikzlibrary{decorations.pathmorphing} 
\usetikzlibrary{fit}					
\usetikzlibrary{backgrounds}	

\tikzset{%
  every neuron/.style={circle,draw,minimum size=1cm},
  neuron missing/.style={draw=none,scale=4,text height=0.333cm,execute at begin node=\color{black}$\vdots$},
}

\title{TArC: Tunisian Arabish Corpus \\
First complete release}

\name{Elisa Gugliotta, Marco Dinarelli} 

\address{Laboratoire d'Informatique de Grenoble (LIG), LIDILEM, ``Sapienza'' University of Rome \\
         Campus Universitaire de Saint-Martin-D'Hères, 38000 Grenoble, \\ Circonvallazione Tiburtina 4, 00185, Rome\\
         \{elisa.gugliotta,marco.dinarelli\}@univ-grenoble-alpes.fr\\}

\abstract{
In this paper we present the final result of a project on Tunisian Arabic encoded in Arabizi, the Latin-based writing system for digital conversations.
The project led to the creation of two integrated and independent resources: a corpus and a NLP tool created to annotate the former with various levels of linguistic information: word classification, transliteration, tokenization, POS-tagging, lemmatization. We discuss our choices in terms of computational and linguistic methodology and the strategies adopted to improve our results. We report on the experiments performed in order to outline our research path. Finally, we explain why we believe in the potential of these resources for both computational and linguistic researches. \\
 }

\begin{document}

\maketitleabstract

\section{Introduction}
\label{sec:Intro}

In this paper we describe the methodology we adopted for building a dialectal Arabic Corpus from scratch. Our motivations for building a corpus from scratch are related to the utility we envisioned with its release, both from a linguistic and a NLP points of view. 
Along with the path identified to achieve our goal, we give the linguistic motivations and describe the computational experiments that guided the choice of our approach, leading us to the final corpus structure.
The corpus is the result of a semi-automatic annotation procedure carried out using an NLP tool based on neural models, that we developed specifically to create the corpus.
Our architecture produces in cascade the different levels of annotation that we decided to have in the corpus.
For many reasons, mentioned below, our approach is \emph{hybrid}.
First of all, the project lies at the intersection of different research fields: Arabic dialectology, corpus linguistics and deep learning.
Secondly, the texts collected in our corpus are in an Arabic dialect. Arabic dialects are notoriously under-resourced linguistic systems.
In particular, the texts collected in our corpus are encoded in a script that, on the one hand represents some phonetic phenomena of Tunisian Arabic (e.g., the article assimilation, unlike the encoding in Arabic script), on the other hand is a writing system.
Such script arose in a diamesically influenced context, namely digital environments.
The encoding we refer to uses the Latin alphabet, as well as some numbers, for Arabic phonemes, without correspondence in the Latin script.
This encoding is known as Franco-Arabic, Arabizi, Arabish \emph{et alias}, depending on the Arabic country. We focused on the system in use for writing Tunisian Arabic.\newline 

The structure of the paper is as follows: we describe the state-of-the-art in section \ref{sec:Related Work}.
In section \ref{sec:Corpus Usefulness} we explain the reasoning behind the planning of our work.
We will describe our corpus building steps in section (\ref{sec:Corpus Structure}).
In section \ref{sec:Architecture} we will present the neural architecture created to annotate our corpus.\footnote{Available at https://gricad-gitlab.univ-grenoble-alpes.fr/dinarelm/tarc-multi-task-system/}
Finally, we will discuss the linguistic-computational methodology to improve our results in section~\ref{sec:Experiments}.
In section~\ref{lemmatization} we describe the procedure to add the lemma annotation level in our corpus.
We conclude the paper in section~\ref{Conclusion}.

\section{Related Work}
\label{sec:Related Work}

Being a Semitic language, Arabic has a complex inflectional and derivational morphology which lead to complex NLP challenges.
Recently, there has been a significant increase in NLP research for morphological processing of both Modern Standard Arabic (MSA) and Dialectal Arabic (DA) \cite{gugliotta-etal-2020-multi}. 
This growth follows two significant independent trends: \textbf{1.} the extension of the application domains of deep learning and \textbf{2.} the rise of social media, leading to the widening of available research data \cite{darwish2021panoramic}.\footnote{The success of word embeddings  \cite{mikolov2013efficient,pennington2014glove} trained on unlabelled data and the resulting enhanced performance on NLP tasks has also helped the growth of interest in Arabic NLP \cite{al2015deep,farha2019mazajak,soliman2017aravec}.}
Among many, a relevant work is the RNN-based model for Arabic morphological disambiguation proposed by \newcite{zalmout2017don}.
The authors exploited the Penn Arabic Treebank (PATB) \cite{maamouri2004penn} and employed LSTM achieving good results.
The increasing availability of easily accessible DA data led to a growing interest in applying Arabic NLP (ANLP) to DA processing
\cite{al2012yadac,bouamor2014multidialectal,bouamor2018madar,diab2010colaba,el2020habibi,gadalla1997callhome,sadat2014automatic,salama2014youdacc,harrat2014building}.

The number of DA corpora increased in the last few years, the majority of them covers the Arabic varieties of Egyptian, Gulf Arabic and the two geographical macro-areas of Levant and Maghreb.
Regarding the last area, one of the most studied variety is Algerian, which, as far as we know, is only represented by three corpora: 
one collected from newspapers and focusing on French code-switching \cite{cotterell2014algerian}; one dealing with Youtube comments \cite{abidi2017calyou}; and one built on an English lexicon automatically translated into Algerian and designed for Sentiment Analysis \cite{guellil2018sentialg}.
There is also a treebank for Algerian, following the Universal Dependencies formalism, containing romanized user-generated contents \cite{seddah2020building}.
There is a good amount of corpora addressing Tunisian Arabic (TA), however often they are not publicly available. 

\textbf{Among the TA corpora} publicly available there are the Tunisian Dialect Corpus Interlocutor (TuDiCoI) \cite{graja2010lexical,graja2013discriminative} and the Spoken TA Corpus (STAC) \cite{zribi2015spoken}.
The latter is morpho-syntactically annotated with a tag set based on the Tunisian-\emph{Al-Khalil} conventions \cite{zribi2013morphological}.
STAC is composed of 42,388 words resulting from transcriptions of audio files from TV channels and radio station. The employed transcription convention is OTTA \cite{zribi2013OTTA}.\footnote{OTTA is a TA dedicated convention for orthographic transcription, oriented to TA phonetics.} 
Another domain-specific corpus is the TA Railway Interaction Corpus (TARIC) \cite{masmoudi2014corpus} built from oral conversations between staff and clients of the Tunisian railway stations.\footnote{TARIC is a task-oriented resource, built for Automatic Speech Recognition (ASR).}
There are also two parallel corpora, one of which is the Parallel Arabic DIalect Corpus (PADIC) \cite{meftouh2015machine,meftouh2018padic}, consisting of 6,400 sentences from six Arabic dialects, with TA among them, aligned at sentence level.
A similar corpus collection was employed by \newcite{bouamor2014multidialectal} for the collection of 2k Egyptian words manually translated into various dialects.
The source texts are part of the Egyptian-English corpus \newcite{zbib2012machine}.
Also the MADAR parallel corpus \cite{bouamor2018madar} has been gathered by translating sentences from English and French into Arabic dialects.
The source texts were collected from the Basic Traveling Expression Corpus (BTEC) \cite{takezawa2007multilingual}. 
A version of MADAR in \emph{CODA} orthography has recently been published.\footnote{Please see the section \ref{subsec:comp_point} for a definition of CODA.} The CODA MADAR Corpus includes 10,000 sentences, together with their original raw version.
The sentences are from MADAR CORPUS-6 which covers the dialects of Beirut, Cairo, Doha, Rabat and TA \cite{eryani2020spelling}.

\textbf{Concerning Arabizi processing}, \newcite{guellil2018sentialg} show that a transliteration task is a required pre-processing stage to decrease the ambiguity of Arabizi, resulting from its lack of spelling conventions.\footnote{In this research, the primary task was the sentiment classification of an Algerian Arabizi corpus.}
In fact, most efforts in Arabizi processing focused on automatic transliteration from Arabizi to Arabic script, such as \newcite{chalabi2012romanized}; \newcite{darwish2014arabizi}; \newcite{al2014automatic}; \newcite{masmoudi2015arabic}; \newcite{younes2016hidden}; \newcite{younes2018sequence}; \newcite{younes2020romanized}.

Studies on other Arabizi features include \newcite{eskander2014foreign}, which focused on foreign words and automatic processing of Arabic social media texts written in Roman script. \newcite{guellil2016arabic} presented an approach for social media dialectal Arabic identification based on supervised methods, using a pre-built bilingual lexicon of 25,086 words proposed by \newcite{guellil2017bilingual} and \newcite{azouaou2017alg}.
A different method is presented in \newcite{younes2018sequence}, where the authors present a sequence-to-sequence model for Tunisian Arabizi-Arabic characters transliteration \cite{sutskever2014sequence}.

As in the case of \cite{younes2020romanized}, most research on automatic processing of Arabizi involves a preliminary phase of, i) corpus collection, ii) model training and testing.
However, creating corpora from scratch is a time and energy demanding practice, and the only available corpora for Tunisian Arabizi are the Electronic Tunisian Dialect (LETD), which gathers 43,222 messages in Latin script collected from the web \cite{younes2014quantitative}; the TLD and TAD  which respectively include 420,897 and 160,418 words in both Latin and Arabic script \cite{younes2015constructing}.\footnote{The latter two corpora are automatically constructed from the web.}
The Tunisian Sentiment Analysis Corpus (TSAC) collects 17,060 Tunisian Facebook comments in Arabic and Latin script, and it is manually annotated with polarity \cite{mdhaffar2017sentiment}. 

\section{Corpus Usefulness}
\label{sec:Corpus Usefulness} 

\subsection{From a Computational Point of View}
\label{subsec:comp_point}

As we can conclude from section~\ref{sec:Related Work}, the available resources for TA and in particular for its Arabizi encoding, are not enough to adequately focus research efforts on the development of automatic TA-Arabizi processing systems. There is indeed a need for data. 
As for the collection of DA corpora, the problem of non-standardized encoding affects all Arabic dialects with different degrees. Whatever type of analysis or computational use of the corpus one wishes to perform, the corpus needs to follow a normalized encoding, as also demonstrated by \cite{guellil2018sentialg}.
This problem was addressed by \newcite{habash2012conventional} by presenting the Conventional Orthography for DA (CODA). This set of guidelines was initially dedicated only to Egyptian dialect, but was later extended to other DA, such as Algerian \cite{saadane2015conventional}, Tunisian \cite{zribi2014conventional}, Maghrebi \cite{turki2016conventional} or Gulf Arabic \cite{khalifa2016large}.
Finally, \newcite{habash2018unified} proposed  \emph{CODA Star}, a common set of orthographic rules focused on features of individual dialects so that to help in creating dialect specific conventions.
CODA Star is the convention we have chosen to make our corpus compatible with other DA corpora. We have provided our corpus with a number of annotation levels that constitute useful information for both computational and linguistic purposes (section~\ref{subsec:ling_point}). Our work can serve as a starting point for numerous and various ANLP research projects thanks to the variety of linguistic annotations it contains and the used methodology. In fact, our corpus was created with a semi-automatic procedure, including a manually check phase (section~\ref{subsec:annotation}). Our code and our data are freely available, the experiments described in  section~\ref{sec:Experiments} are thus reproducible.\footnote{The code is available at: https://gricad-gitlab.univ-grenoble-alpes.fr/dinarelm/tarc-multi-task-system, the Corpus is available at: https://github.com/eligugliotta/tarc.}     

The neural architecture was designed for our main goal: building the multi-level annotated Tunisian Arabish Corpus (TArC), and it could be used for extending our project in the future. Moreover, it could be possible to adapt the same tool to other DA texts, in whatever encoding they are written in, creating other annotated twin corpora covering other Arabic dialects.

\subsection{From a Linguistic Point of View}
\label{subsec:ling_point}

As the TA lacks resources for automatic processing, it also lacks resources for its linguistic study.   
There are very few resources available for this purpose.
Regarding TA self-learning there are a number of scientific studies that describe its feature (e.g.,  \newcite{gibson2011tunis}; \newcite{baccouche2011tunisia}; \newcite{veronika2014hilal}; \newcite{mion2017propos}), some grammar  manuals (e.g., \newcite{abdelkader1977peace}; \newcite{stumme1896grammatik}; \newcite{singer1984grammatik}), digital corpora questionable via interface, such as the \emph{Tunisian Arabic Corpus} \cite{mcneil2018tunisian} and the corpus released in the context of the \emph{TUNICO} project \cite{moerth2014laying,Moerth2017linking}.
Some dictionaries are also available through web interfaces (e.g., that of the \emph{TUNICO} project), or downloadable (e.g., \emph{Le Karmous} \cite{abdellatif2010dictionnaire}, as well as some ontologies \cite{karmani2015aebwordnet,karmani2014building,moussa2015tunisian}).
However, as much as these tools are unquestionably useful in supporting research, TA still remains rather uncovered either in terms of global materials, or in terms of support tools for linguistic analyses.
This is the reason why our corpus has been annotated with various linguistic information and metadata.
The amount of data is certainly one of the most important elements in NLP research, but the accuracy of the data and the levels of linguistic information are no less important and are certainly necessary for linguistic research. For this reason, we chose to concentrate on a reduced amount of data in order to focus on a methodology that could meet linguistic needs.\footnote{However, we plan to increase the size of the corpus by exploiting TArC as a gold standard.}
In order to enable different types of research through our work, we decided to provide as many levels of information as possible on the selected texts. 
For instance, we chose to collect texts from different Digital Networked Writing (DNW) environments, such as forums, blogs and social networks. In this way, TArC allows analyses that correlate language and spelling adopted by users according to the media platform employed. 
In order to be able to carry out comparative analyses among the different `textual genres', we decided to limit our intervention on the original data as much as possible. This means that we have not removed any element from the texts, i. e. punctuation, symbols, typos, nor all those para-textual elements typical of the DNW.\footnote{E.g., emoticons or smileys, which in our word classification system converge in the \emph{emotag} class (section \ref{2nd_phase}).} The subdivision of paragraphs into sentences respected textual semantics as much as possible, relying on both syntax and end-of-clause punctuation (i.e. '.', '!', '?'). All these measures motivated by linguistic reasons involved compromises with the typical NLP \emph{modus operandi}. For example, we chose to treat the different `textual genres' separately, in different blocks so as not to compromise the natural order of communicative exchanges or sentences in paragraphs. In doing so, however, we sacrificed the homogeneous distribution of `textual genres' in the annotation blocks. This, on the other hand, allowed us to observe different model learning behaviors at different stages of text annotation. In this regard, in section~\ref{sec:Experiments}, we discuss how we trace these different behaviors to the different nature of the texts processed at each annotation phase. Finally, these observations on the architecture behavior supported our linguistic analyses by confirming some structural differences between `text genres'. 
Along with analyses of the different `textual genres', our data also allow to observe some typical traits of Tunisian Arabic and of its spontaneous writing system \cite{gugliottaforthcoming}.\footnote{We have already carried out preliminary analyses of this type \cite{gugliottaforthcoming}.}
The data extracted from the lyrics of rap songs encoded in Arabizi allow comparison with their encoding in Arabic characters, as we have chosen songs that we know are widespread in both writing systems. Thanks to the fact that, in the data collection phase (section \ref{subsec:data_collection}), we collected both the texts and their metadata, it is possible, for example, to carry out diachronic, diatopic, diastratic and generally sociolinguistic analyses on TArC. Thanks to the annotation levels provided within our corpus, all these linguistic analyses can concern the orthographic, morphological and syntactic levels.

\section{The Corpus}\label{sec:Corpus Structure}

\subsection{Data collection}\label{subsec:data_collection}
Considering the nature of the Arabizi encoding, we primarily identified three sources of digital conversations in Tunisian Arabic to collect the data we wanted to include in our corpus.
These sources of written texts are conversations on social networks, forums, and texts extracted from blogs. We were also interested in extracting some musical texts for the purpose of comparison between the two writing systems, considering that for the most popular songs it is possible to find both encodings. So we also extracted some lyrics of rap music (the most popular genre among young Tunisians). Our goal was to collect texts of varying lengths, from medium to long, so that they contain as much context as possible and are rich in linguistic information.
Thus, concerning social network texts, we ruled out the possibility of using Twitter and with it the possibility of automatic filtering texts by location. In order to quickly identify Tunisian texts, we had to organize our crawling based on a keyword search so as to guide the automatic online texts searching and URLs list creating. Thus, we adopted a methodological approach similar to that one outlined in \newcite{saadane2018automatic}. 
With the aim of building a corpus as representative as possible of the linguistic system, we considered useful to identify wide thematic categories that could represent the most common topics of online daily conversations.
In this regard, two instruments with a similar thematic organisation have been employed, i.e., ‘A Frequency Dictionary of Arabic’ and in particular its ‘Thematic Vocabulary List’ (TVL) \cite{buckwalter2014frequency} and the ‘Loanword Typology Meaning List’, which is a list of 1460 meanings (LTML) \cite{haspelmath2009loanword}.
By aiming to prevent introducing relevant query biases, it was decided to avoid the use of category names in the query, but to generate a range of keywords \cite{schafer2013web}. Therefore, each category was associated with a set of keywords in Arabizi belonging to the basic Tunisian vocabulary. Three meanings for each semantic category was found to be enough to obtain a sufficient number of keywords and URLs for each category. E.g., for the category ‘family', the meanings: ‘child', ‘marriage', ‘divorce' were associated with all their TA variants, resulting in an average of 10 keywords for each macro-category.\footnote{This phase is more deeply described in  \newcite{gugliotta2020tarc}.}
After manually checking the URLs list, in order to ensure the compatibility of the identified pages with the project aims, the following stage was the automatic scraping of the selected pages. In this way, we built a corpus that covers the basic lexicon terms in a balanced way. Moreover, this methodology avoids any bias typically introduced by manual queries based on thematic keywords \cite{rinke2022expert}.


Some quantative information about the extracted data, are reported in table~\ref{tab:ArabiziStats}. 

\begin{table}[ht]
        \centering
\scalebox{0.68}{
        \begin{tabular}{|l|rr|rr|rr|rr|}
        \hline
          & \multicolumn{1}{c}{\emph{Sentences}} & \multicolumn{3}{|c}{\emph{Words}}
          & \multicolumn{1}{c|}{}& \multicolumn{1}{c|}{\emph{Avg sentence len.}}\\
        \hline
        Total
          &\multicolumn{1}{c}{4,797} &\multicolumn{3}{|c}{43,327}&\multicolumn{1}{c|}{} & \multicolumn{1}{c|}{9.0}\\
        \hline
        \hline
        \emph{forum} & \multicolumn{1}{c|}{755}   & \multicolumn{1}{c}{} & \multicolumn{1}{c}{11,909} &
        \multicolumn{1}{c}{} & \multicolumn{1}{c|}{} & \multicolumn{1}{c|}{15.8} \\
        \emph{social} &  \multicolumn{1}{c|}{3,162}   & \multicolumn{1}{c}{} & \multicolumn{1}{c}{16,056} &
        \multicolumn{1}{c}{} & \multicolumn{1}{c|}{} & \multicolumn{1}{c|}{5.1} \\
        \emph{blog} &  \multicolumn{1}{c|}{366}  & \multicolumn{1}{c}{} & \multicolumn{1}{c}{6,671} &
        \multicolumn{1}{c}{} & \multicolumn{1}{c|}{} & \multicolumn{1}{c|}{18.2}\\
        \emph{rap} &  \multicolumn{1}{c|}{514}  & \multicolumn{1}{c}{} & \multicolumn{1}{c}{8,691} &
        \multicolumn{1}{c}{} & \multicolumn{1}{c|}{} & \multicolumn{1}{c|}{16.9} \\
        \hline
        \end{tabular}
}
        \caption{Statistics of our corpus}
    \label{tab:ArabiziStats}
\end{table}

\subsection{Semi-Automatic Annotation}\label{subsec:annotation}

\begin{table*}[t]
\begin{center}
\scalebox{0.68}{
    \begin{tabular}{|l|c|cccc|}

        \hline
        & \textbf{Train. tokens} & \multicolumn{4}{c|}{\textbf{Tasks (Accuracy)}} \\
        \textbf{Step} & & \textbf{Cl} & \textbf{Ar} & \textbf{Tk} & \textbf{POS} \\
        \hline
        \hline
        \multicolumn{6}{|c|}{Corpus: MADAR} \\
        \hline
        Step0 & 12,391  & 99.83   &   -   &   88.83   &   72.71 \\
        Step0$_{\text{complete}}^{(*)}$ & 12,391  & 99.58   &   76.77   &   74.83   &   67.59 \\
        \hline
        \hline
        \multicolumn{6}{|c|}{Corpus: MADAR+TArC} \\
        \hline
        Step1 & 17,261 (4,870)  & 92.69   &   -   &   77.66   &   59.56 \\
        Step2 & 22,173 (9,780)  & 97.21   &   -   &   87.53   &   74.30 \\
        Step3 & 27,270 (14,870)  & 96.69   &    -   &   91.47   &   76.38 \\
        \hline
        \hline
        \multicolumn{6}{|c|}{Corpus: TArC} \\
        \hline
        Step4 & 22,150  &   96.83 &   75.30 &   73.38 &   69.76 \\
         Step5 & 27,435  &   97.17 &   75.08 &   73.07 &   66.24 \\
         \hline
         Step4$_{\text{smart-init}}$ & 22,150  &   95.91 &   76.55 &   74.96 &   72.57 \\
         Step5$_{\text{smart-init}}$ & 27,435  &   97.08 &   77.83 &   75.69 &   69.76 \\
         \hline
         \hline
        \multicolumn{6}{|c|}{Corpus: MADAR$_{\text{Arabizi}}$+TArC} \\
        \hline
         Step4$_{\text{concat}}$$^{(*)}$ & 34,541 (22,150)  &   96.59    &   78.94 &   77.38 &   74.54 \\
         Step4$_{\text{reloaded}}$$^{(*)}$ & 34,541 (22,150)  &   96.38    &   79.72 &   77.88 &   73.69 \\
        \hline
        \hline
        Step6$_{\text{concat}}$$^{(*)}$ & 46,197 (33,806)  &   96.45    &   79.97 &   77.81 &   70.33 \\
        Step6$_{\text{concat}}$$^{(*)}$ \textbf{fix} & 46,197 (33,806)  &   97.63    &   83.29 &   81.94 &   81.02 \\
        \hline
        \hline
        Final-Step$_{\text{concat}}$$^{(*)}$ lstm & 42,895 (30,504)  &   98.56    &   82.98 &   81.84 &   82.84 \\
        Final-Step$_{\text{concat}}$$^{(*)}$ transformer & 42,895 (30,504)  &   95.99    &   75.37 &   74.34 &   71.30 \\
        \hline
        Final-Step$_{\text{concat}}$$^{(*)}$ input:Ar lstm & 42,895 (30,504)  &   98.67    &   - &   96.78 &   86.31 \\
        Final-Step$_{\text{concat}}$$^{(*)}$ input:Ar transformer & 42,895 (30,504)  &   99.95    &   - &   95.93 &   82.49 \\
        \hline
        
        \hline
    \end{tabular}
}
\end{center}
\caption{Summary of results, in terms of accuracy, obtained on the TArC data at the different steps of the iterative procedure for semi-automatic annotation of the corpus. The tasks are indicated with \textbf{Cl} for classification, \textbf{Ar} for Arabic script encoding, \textbf{Tk} for tokenization, and \textbf{POS} for POS tagging. $^{(*)}$ indicates results obtained with the MADAR data translated into Arabizi.}
\label{tab:MADATarc_AccIterativeResults}
\end{table*}

In order to make the collection of TArC as fast and as easy as possible for human
annotators, deep learning techniques have been employed to implement a semi-automatic
annotation procedure \cite{gugliotta2020tarc}.
In particular the multi-task sequence-to-sequence neural architecture described in \newcite{gugliotta-etal-2020-multi} has been used.
Such a system takes one or more input as sequences, and generates one or more outputs as sequences.
The number of outputs is dynamically and automatically detected by the system based on the data format.
The same system has been thus used for different phases of the annotation procedure, where a different number of annotation levels was available (see below).
A high level schema of the multi-task system is given in the figure~\ref{fig:multitasksystem}, where the system is instantiated to generate all the annotation levels of the corpus, taking Arabizi text as input.

The iterative semi-automatic procedure adopted to collect the TArC corpus consists in splitting the data to be annotated in blocks of roughly the same size and then:
\begin{enumerate}[noitemsep,nolistsep]
    \item Annotating automatically a block of data with a model;
    \item Correcting manually the automatic annotation;
    \item Adding the new annotated block of data to the training data of the model;
    \item Training a new model;
    \item Restarting from step 1 with a new block of data.
\end{enumerate}

The TArC data have been split into seven blocks of roughly 6,000
tokens, requiring thus seven steps of the procedure to annotate the whole corpus.
Concerning the annotator, he is a non-native speaker with a background in Tunisian Arabic.\footnote{Having first specialized in Standard Arabic and having been living in Tunisia for a few years. In fact, he attended the \emph{IBLV} and the \emph{WALI} in Tunis, focusing on Tunisian Arabic.} He could rely on the tools mentioned in section \ref{subsec:ling_point} and the constant feedback from native speakers.

\subsubsection{First Annotation Phase}\label{1st_phase}

In order to facilitate correspondence with existing tools and studies concerning dialectal Arabic processing, we considered essential to provide our corpus with a  normalized encoding level.
Arabizi is indeed a spontaneous writing-system and does allow different encodings for a given Tunisian lexeme. 
In order to achieve this normalization, we decided to transliterate the texts in a conventional orthography created \emph{ad hoc} for Arabic dialects processing, the CODA Star orthography \cite{habash2018unified}.
Since at first no annotated data existed, in order to start the semi-automatic procedure some data must be manually annotated to train the model from scratch.
A block of the TArC corpus has been manually annotated with the Tunisian Arabic transliteration of Arabizi tokens. 
We must point out that Arabizi encoding, being a Latin-based script, allows a consistent employ of code-switching and transfers. In case of Tunisian Arabic, these elements are mostly coming from French language as can be seen in table \ref{tab:corpus}. 
Considering the linguistic aims of our work, transliterating French tokens into CODA would have generated confusion in the data, namely hiding word etymology, and that would have made it difficult to perform linguistic analyses on our corpus.\footnote{For example, the hypothetical transliteration of the French insertion `ma grossesse' in the \ref{tab:corpus} table could have been 
\RL{ما ڤروساس},
where the transcription of the possessive adjective overlaps with that of the Tunisian particle (\RL{ما}) as well as with the noun used to denote `water'. }
Furthermore, the correspondence between the phonological and orthographic levels of Tunisian and European languages is necessarily asymmetric. These asymmetries would have definitely resulted in noise for an automatic transliteration model. 
Aware of needing a better solution (addressed in the second phase~\ref{2nd_phase}), we decided to work only with non-code-switched data, manually tagging the other tokens as \emph{foreign}. In this way, our first block of data was reduced from roughly 6,000 tokens to 5,000. The same has been done on two more blocks of data. The automatic annotation accuracy at the end of the first phase was roughly 65\%. 
For more information regarding this phase we refer to \newcite{gugliotta2020tarc}. 
Since the Arabizi transliteration into CODA, shown in table \ref{tab:corpus}, is the most difficult and ambiguous phase due to the spontaneous nature of Arabizi \cite{gugliottaforthcoming}, before going on with the annotation of the other data blocks, it has been decided to implement a more comprehensive strategy. The latter consists of the second phase, and can be resumed in the following points: 
\begin{enumerate}[noitemsep,nolistsep]
    \item Adding an Arabizi token classification level before the transliteration level;
    \item Improving performance in the transliteration task exploiting the information included in the other annotation layers we wanted to perform.  
\end{enumerate}

Hence, the intuition to continue the corpus annotation, instead of through separate modules, using a multi-task architecture that could allow the different modules to benefit from shared information. This is what we refer to as the second phase, outlined in the next section. 






\subsubsection{Second Annotation Phase}\label{2nd_phase} 

The levels of annotation provided in our corpus are classification of tokens in three categories (\emph{arabizi}, \emph{foreign} and \emph{emotag}), transliteration into Arabic conventional script (CODA), word reduction in its morphemes (tokenization), Part-of-Speech tagging and lemmatization.\footnote{The POS-tagging formalism employed  includes 184 tags and it is an adaptation of the Buckwalter tag set for MSA. We followed the guidelines of the PATB \cite{maamouri2004patb2}.}
\begin{table}[htbp]\small \label{tab:corpus}
\begin{center}
\scalebox{0.68}{
\begin{tabularx}{\columnwidth}{cccc}
     \hline
     CODA & Tokeniz. & POS & Lemma\\
     \hline
     \hline

     \RL{انا} & \RL{انا} &  \small{PRON\_1S}& \RL{هو} \\
     
     \RL{بعد} & \RL{بعد} & \small{ADV} & \RL{بعد} \\
     
     \emph{ma} &  foreign  & foreign & foreign \\
    
     \emph{grossesse}  & foreign &
     foreign & foreign\\

     \RL{حوايجي} &  \RL{حوايج+ي} & \small{NOUN+} & \RL{حوايج}\\
     & & \small{POSS\_PRON\_1S}  \\
     
    \RL{ال} & \RL{ال} & \small{DET } & \RL{ال}  \\
     
     \RL{قدم} & \RL{قدم} & \small{ADJ } & \RL{قديم}  \\
     
     \RL{ال} & \RL{ال} & \small{DET } & \RL{ال}  \\
     
     \RL{كلّهم} &  \RL{كلّ+هم} & \small{NOUN\_QUANT+}
     & \RL{كلّ}\\
     & & \small{PRON\_3P} \\
     
     \RL{ولّاوا} & \RL{ولّاوا} & \small{PV-PVSUFF\_} & \RL{ولّى}\\
     & & \small{SUBJ:3P} \\
     
     \emph{motivation} & foreign & foreign & foreign \\
     \hline
  \end{tabularx}
}
    \caption{An excerpt of TArC annotation levels of the Arabizi sentence \emph{ena ba3d ma grossesse houayji el kdom el kollehom waleou motivation} (`after my pregnancy all my previous clothes became my motivation').}
    \label{TARCtable}
     \end{center}
    \end{table}

Regarding the procedure employed to classify our data we refer to \newcite{gugliotta-etal-2020-multi}. 
Concerning the other levels, these have been performed in the following way. First of all, in order to extend the amount of Tunisian Arabic data,
it has been decided to exploit a morpho-syntactically well-formed corpus.
Thus, 2,000 Tunisian Arabic sentences (roughly 12,400 tokens) from the MADAR parallel corpus \cite{bouamor2018madar} 
have been annotated with the annotation levels planned for our corpus, namely classification, tokenization and POS tags, applying the semi-automatic procedure described in \ref{subsec:annotation}.\footnote{Due to time constraints, we started working on lemmatization level in recent times. The inherent procedure is described in section \ref{lemmatization}.} 
After annotating the MADAR corpus with the  mentioned levels, the semi-automatic procedure for annotating the whole TArC corpus started.\footnote{We decided to use the MADAR corpus instead of the Tunisian corpora mentioned in section \ref{sec:Related Work}, as we considered it more compatible for our corpus-building goals and future projects, namely to extend towards other Maghrebi varieties.} Results in terms of accuracy at each step are reported in table \ref{tab:MADATarc_AccIterativeResults}, and are described in details in section~\ref{sec:Experiments}.

\section{The Architecture}\label{sec:Architecture}

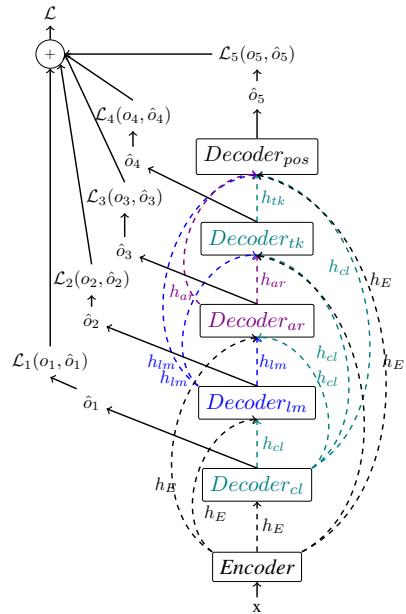
\begin{figure}[!ht]
\center
\scalebox{0.68}{
\begin{tikzpicture}[scale = 0.8]
	\begin{scope}[local bounding box=net]
	
	\node (o) at (0,11.5) {$\hat{o}_{5}$};

	\node[draw, rectangle, rounded corners=1pt, scale=1.2] (d5) at (0,10.0) {\emph{$Decoder_{pos}$}};

	\node[draw, rectangle, rounded corners=1pt, scale=1.2, text=teal] (d4) at (0,8.0) {\emph{$Decoder_{tk}$}};
	
	\node[draw, rectangle, rounded corners=1pt, scale=1.2, text=violet] (d3) at (0,6.0) {\emph{$Decoder_{ar}$}};
	
	\node[draw, rectangle, rounded corners=1pt, scale=1.2, text=blue] (d2) at (0,4.0) {\emph{$Decoder_{lm}$}};
	
	\node[draw, rectangle, rounded corners=1pt, scale=1.2, text=teal] (d1) at (0,2.0) {\emph{$Decoder_{cl}$}};
	
	\node[draw, rectangle, rounded corners=1pt, scale=1.2] (e) at (0,0.0) {\emph{Encoder}}; 
	
    \node (i) at (0,-1.0) {x};
    
    \draw[->, thick] (i) -- (e);
    \draw[->, thick, dashed] (e) -- (d1);
    \draw[teal, ->, thick, dashed] (d1) -- (d2);
    \draw[blue, ->, thick, dashed] (d2) -- (d3);
    \draw[violet, ->, thick, dashed] (d3) -- (d4);
    \draw[teal, ->, thick, dashed] (d4) -- (d5);
    \draw[->, thick] (d5) -- (o);

    \draw[teal, ->, thick, dashed] (d1.north east) to[bend right=70] (d3.south); 
    \draw[teal, ->, thick, dashed] (d1.north east) to[bend right=70] (d4.south); 
    \draw[teal, ->, thick, dashed] (d1.north east) to[bend right=70] (d5.south); 
    
    \draw[blue, ->, thick, dashed] (d2.north west) to[bend left=70] (d4.south); 
    \draw[blue, ->, thick, dashed] (d2.north west) to[bend left=70] (d5.south); 

    \draw[violet, ->, thick, dashed] (d3.north west) to[bend left=70] (d5.south); 
    
    \draw[->, thick, dashed] (e.north east) to[bend right=70] (d4.south); 
    \draw[->, thick, dashed] (e.north east) to[bend right=70] (d5.south); 
    \draw[->, thick, dashed] (e.north west) to[bend left=70] (d3.south);  
    \draw[->, thick, dashed] (e.north west) to[bend left=70] (d2.south);

    \node (o1) at (-4.0,4.0) {$\hat{o}_{1}$};
    \node (o2) at (-4.0,6.0) {$\hat{o}_{2}$};
    \node (o3) at (-3.2,7.7) {$\hat{o}_{3}$};
    \node (o4) at (-3.0,9.9) {$\hat{o}_{4}$};
    
    \node (L1) at (-5.0,5.0) {$\mathcal{L}_1(o_1, \hat{o}_1)$};
    \node (L2) at (-4.0,7.0) {$\mathcal{L}_2(o_2, \hat{o}_2)$};
    \node (L3) at (-3.2,9.0) {$\mathcal{L}_3(o_3, \hat{o}_3)$};
    \node (L4) at (-3.0,11.0) {$\mathcal{L}_4(o_4, \hat{o}_4)$};
    \node (L5) at (-0.0,12.5) {$\mathcal{L}_5(o_5, \hat{o}_5)$};
    \node[draw, circle, scale=0.8] (L) at (-5.0,12.5) {$+$};
    \node (LL) at (-5.0,13.5) {$\mathcal{L}$};
    
    \draw[->, thick] (d1.north) -- (o1);
    \draw[->, thick] (d2.north) -- (o2);
    \draw[->, thick] (d3.north) -- (o3);
    \draw[->, thick] (d4.north) -- (o4);
    
    \draw[->, thick] (o1.north west) -- (L1.south);
    \draw[->, thick] (o2.north) -- (L2.south);
    \draw[->, thick] (o3.north) -- (L3.south);
    \draw[->, thick] (o4.north) -- (L4.south);
    \draw[->, thick] (o.north) -- (L5.south);
    \draw[->, thick] (L1.north) -- (L.south);
    \draw[->, thick] (L2.north) -- (L.south east);
    \draw[->, thick] (L3.north) -- (L.east);
    \draw[->, thick] (L4.north) -- (L.east);
    \draw[->, thick] (L5.west) -- (L.east);
    \draw[->, thick] (L.north) -- (LL.south);

    \node (hE1) at (0.4,1.0) {$h_E$}; 
    \node (hE2) at (-1.0,1.3) {$h_E$}; 
    \node (hE3) at (-2.2,2.0) {$h_E$}; 
    \node (hE4) at (3.3,5) {$h_E$}; 
    \node (hE5) at (3.0,7) {$h_E$}; 
    
    \node (h11) [text=teal] at (0.4,3.0) {$h_{cl}$}; 
    
    \node (h12) [text=teal] at (1.7,4.6) {$h_{cl}$}; 
    \node (h13) [text=teal] at (1.7,5.2) {$h_{cl}$}; 
    \node (h14) [text=teal] at (2,7.3) {$h_{cl}$}; 
    
    \node (h15) [text=blue] at (-2.0,4.6) {$h_{lm}$}; 
    \node (h16) [text=blue] at (-2.3,5.0) {$h_{lm}$}; 
    
    \node (h18) [text=violet] at (-1.8,6.7) {$h_{ar}$};

    \node (h2) [text=blue] at (0.4,5.0) {$h_{lm}$};
    \node (h3) [text=violet] at (0.4,7.0) {$h_{ar}$};
    \node (h4) [text=teal] at (0.4,9.0) {$h_{tk}$};

    \end{scope}

\end{tikzpicture}
}
    \caption{\emph{A high-level schema of our architecture}}\label{fig:multitasksystem}
\end{figure}

As previously mentioned, considering the correlation between the different annotation levels designed for the TArC corpus, we had the intuition to produce these levels with a neural Multi-Task Architecture (MTA) rather than generating them one by one with mono-task models. 
Indeed, a neural architecture learns to factorize information across tasks, even when employing different modules for different tasks, which are learned jointly and interdependently. Thanks to these properties, the predictions at different levels should improve their individual performance.
Based on our intuition about the need to filter Arabizi data through a level of classification, we put this level as first in the MTA, followed respectively by the modules dedicated to lemmatization, transliteration in Arabic script (CODA), tokenization and POS-tagging.
Each of these modules corresponds to a \emph{Decoder} of our neural architecture in figure \ref{fig:multitasksystem}.
Thus, we have the \emph{$Decoder_{cl}$} for classification task,  \emph{$Decoder_{lm}$} for lemmatization,   \emph{$Decoder_{ar}$} for transliteration into Arabic script, \emph{$Decoder_{tk}$} for tokenization and \emph{$Decoder_{pos}$} for PoS tagging.
The input ($x$), consisting of the Arabizi text embedding, is converted into context-aware hidden representations by the \emph{$Encoder$}. Each decoder is provided with a number of attention mechanisms equivalent to the number of previous modules (including the encoder), it receives thus as input the encoder's hidden state (\emph{$h_E$}) together with the hidden state of each previous decoder.
Each decoder returns its predicted output ($\hat{o}_i$ for $i = 1 \dots 5$), which is used to learn the corresponding task by computing a loss comparing the predicted to the expected output (i.e. {$\mathcal{L}_i(o_i, \hat{o}_i)$}). The whole architecture, and thus all the tasks, is learned end-to-end by computing a global loss $\mathcal{L}$ via the sum of each individual loss ($\mathcal{L} = \sum_{i=1}^{5} \mathcal{L}_i(o_i, \hat{o}_i)$), represented with the circled $+$ symbol at the upper part of the figure~\ref{fig:multitasksystem}.

\section{Experiments}\label{sec:Experiments}

\begin{table*}[t]
\begin{center}
\scalebox{0.68}{
    \begin{tabular}{|l|c|ccccc|}

        \hline
        & \textbf{Train. tokens} & \multicolumn{5}{c|}{\textbf{Tasks (Accuracy)}} \\
        \textbf{Step} & & \textbf{Cl} & \textbf{Ar} & \textbf{Tk} & \textbf{POS} & \textbf{Lm} \\
        \hline
        \hline
        \multicolumn{7}{|c|}{Corpus: MADAR$_{Arabizi}$+TArC} \\
        \hline
        Step1 & 17,509 (5,118)  & 98.61 &   73.61   &   73.15   & 73.92 & 72.22 \\
        Step2 & 22,272 (9,881)  & 97.33   &   79.10   &   77.53   &   78.82 & 75.14 \\
        Step3 & 27,399 (15,008)  & 98.31   &   80.69   &   79.81   &   80.38 & 79.00 \\
        Step4 & 33,069 (20,678)  & 99.13   &   81.77   &   80.94   &   82.30 & 80.36 \\
        Step5 & 38,681 (26,290) & 98.72 & 85.79 & 84.89 & 85.44 & 83.69 \\
        Step6 & 44,792 (32,401)  & 97.13 & 85.96 & 84.81 & 83.11 & 84.38 \\
        Final Step global-split & 42,559 (30,168)  & 97.14 & 82.34 & 81.45 & 80.95 & 80.48 \\
        Final Step genre-split & 42,559 (30,168)  & 98.47 & 82.93 & 81.77 & 80.33 & 81.40 \\
        \hline
        \hline
        \textbf{Step} & \textbf{Train. tokens} & \textbf{Cl} & \textbf{Lm} & \textbf{Ar} & \textbf{Tk} & \textbf{POS} \\
        \hline
        Final Step 2xlstm & 42,559 (30,168) & 98.42 & 81.81 & 82.65 & 81.58 & 81.60 \\
        Final Step 3xtransformer & 42,559 (30,168)  & 96.48 & 68.89 & 69.72 & 68.18 & 68.37 \\
        \hline
        Final Step 2xlstm input:Ar & 42,559 (30,168) & 98.77 & 92.40 & - & 96.74 & 85.90 \\
        Final Step 3xtransformer input:Ar & 42,559 (30,168) & 96.91 & 83.10 & - & 93.43 & 74.09 \\
        \hline
    \end{tabular}
}
\end{center}
\caption{Summary of results, in terms of accuracy, for the semi-automatic procedure for TArC lemmatization. \textbf{Lm} stays for Lemma, the other annotation levels are like in table~\ref{tab:MADATarc_AccIterativeResults}.}
\label{tab:TarcLemmatization}
\end{table*}

In table~\ref{tab:MADATarc_AccIterativeResults} we report results for all annotation levels except lemmatization, which has been added afterwards and will be described in section~\ref{subsec:lemmatization}.
We note that results up to the block marked with \textit{Corpus: TArC} included, have been already described in \newcite{gugliotta-etal-2020-multi}, and in \newcite{gugliottaforthcoming}.
We report them here for completeness, but we give only a short description.

In the table~\ref{tab:MADATarc_AccIterativeResults}, \textbf{Train. tokens} indicates the number of tokens used for training the model at each step, in parenthesis we show the number of tokens from the TArC corpus (the others are from the MADAR corpus).
Accuracy for the four annotation levels are reported on the rightmost part of the
table.
The line corresponding to Step0$_{\text{complete}}^{(*)}$ describes an experiment performed in a later time, with respect to the other steps reported in the first lines of the table~\ref{tab:MADATarc_AccIterativeResults}. Such an experiment will be explained later on in this section.

In order to facilitate the understanding of the different annotation steps of the procedure, we note that at step $i$, $i$ blocks of TArC data are used for training the model.
At step 0 thus, only MADAR is used as training data to annotate the first block of TArC data. This is the bootstrap step to start the procedure.
As reported in table~\ref{tab:MADATarc_AccIterativeResults}, the incremental procedure has been applied up to step 3 for annotating data with Classification, Tokenization, and POS tags using Tunisian Arabic written in CODA as input.
We note that for these three annotation steps, accuracies are relatively high (at best 91.47\% and 76.38\% at step 3 for Tokenization and POS tagging, respectively).

In step 4 and 5 of the incremental annotation procedure, Arabizi has been used as input to the system, since only the first three data blocks were annotated with tokens transliterated in CODA.
For these two steps thus, not only the system has less training data, but also an additional decoder is instantiated to predict the CODA annotation level.
This means also that the model has more parameters to train.
All of that translates overall into a drop in performance when using Arabizi as input to the system (accuracies are 77.83\%, 75.69\%, 69.76\% for transliteration into CODA, Tokenization and POS tagging, respectively, at step 5).

In order to further reduce data sparsity and ambiguity, the latter especially related to predicting Arabic script from Arabizi, MADAR data have been also annotated with Arabizi encoding level.
This has been performed once again with a semi-automatic procedure.
The step 0 has been performed again, this time with Arabizi as input to predict the other 4 levels, allowing to compare results with the very first step 0 of the annotation procedure.
Such an experiment is indicated with Step0$_{\text{complete}}^{(*)}$ in the table~\ref{tab:MADATarc_AccIterativeResults} (meaning that MADAR data have now all the annotation levels).
We note that, once again predicting all the information levels from Arabizi leads to an overall drop in performances.
Since MADAR is composed of morpho-syntactically well-formed text, this confirms that predicting CODA level from Arabizi is a difficult task, and the most difficult among those performed in this multi-task setting.
As we explain in a forthcoming work (to appear), while concatenating the MADAR data to the TArC data provides similar performances with respect to initializing the model with one pre-trained on MADAR data only (Step4$_{\text{concat}}$ vs. Step4$_{\text{reloaded}}$ in the table), the former is slightly more accurate on POS tagging and doesn't require to pre-train a model, allowing to save resources.
The concatenation strategy has thus been chosen for the remainder of the experiments.
We refer the reader to \cite{gugliotta2020tarc,gugliotta-etal-2020-multi} for more details on this part of the experiments and on experimental settings.

The step 6 is the last step for annotating the data block 7 of the TArC corpus.
Results on this step are similar to those of previous steps, with a small drop on POS tagging. We attribute this to the fact that the block 7 is of a different genre with respect to previous blocks.

At this point of the annotation procedure we decided to update our multi-task system adding the Transformer model \cite{46201}. This allowed to find a weight initialization problem in the system.\footnote{Basically parameters were initialized with the default of Pytorch (https://pytorch.org/docs/stable/index.html), which we replaced with the more effective \textit{Xavier initialization} \cite{PracticalRecommendations:Bengio:2012}.} Solving this problem not only allowed to have a working system with both LSTM and Transformer \cite{Hochreiter-1997-LSTM,46201}, but also improved drastically the performances of the system when using LSTM.
This can be seen in table~\ref{tab:MADATarc_AccIterativeResults} comparing the lines Step6$_{\text{concat}}$ and Step6$_{\text{concat}}$ \textbf{fix}.
Despite the huge change in performances, we decided to not perform again the previous experiments. This first of all because the experiments were needed to pre-annotate data, while pre-annotation and manual correction up to this step had already been performed. Second in order to save resources, and third because even with the performance change the main message stays the same: the most difficult and ambiguous task is still the transliteration of Arabizi tokens into CODA script. This is confirmed by the fact that performance drops (relatively) very little from Ar to POS prediction (83.29\% to 81.02\%).
The remainder of the experiments has been performed with the corrected system.

Once the whole corpus has been annotated with all the annotation levels, we split the data into training, development and test splits after a random shuffle at sentence level. This has been performed by splitting separately each text genre with 70/15/15 ratios for the three splits, respectively, and then concatenating the corresponding splits of all genres.
Experiments with this split are reported in table~\ref{tab:MADATarc_AccIterativeResults} with Final-Step$_{\text{concat}}$, and have been performed with both LSTM and Transformer for comparison. In addition, we performed also experiments using Tunisian in CODA script as input (input:Ar) to predict the other levels (except for Arabizi). Similar experiments have been performed for annotation steps from 1 to 3, and they could be useful for automatically annotating more data to be used in a similar strategy as \emph{back translation} \cite{sennrich-etal-2016-improving}.

\subsection{Lemmatization}\label{lemmatization}
\label{subsec:lemmatization}

Though the lemmatization annotation level was planned since the begin of our project, it was the last level we produced for practical reasons.
Despite its unarguable usefulness, transliteration in CODA, tokenization and POS tagging were the most crucial annotation layers, in particular for linguistic analyses.
However, even lemmatization represents a tool of fundamental importance both for linguistic analysis and automatic data processing \cite{zalmout2019joint}. 
The lemmatization level was produced using the same semi-automatic annotation procedure used for the other levels.
Again, we exploited data from the MADAR corpus, which were semi-automatically lemmatized, using a first block of manually lemmatized TArC data for bootstrapping the annotation procedure.
The TArC lemmas are encoded in Arabic CODA Star orthography, also used for the transliteration level.

The results for the lemmatization procedure are reported in table~\ref{tab:TarcLemmatization}, with the same format as previous results. We note that results obtained with LSTMs up to step 6 are substantially better than those in table~\ref{tab:MADATarc_AccIterativeResults}. This is due to the use of the corrected system for all steps (see previous section), but also to the presence of the lemmas. Indeed, by comparing the step 6 in table~\ref{tab:MADATarc_AccIterativeResults} and in table~\ref{tab:TarcLemmatization}, all performed with the corrected system, we can see that results are better when lemmas are used, confirming our intuition on the usefulness of this annotation level.

We note also that, overall, Transformers are substantially less effective than LSTMs on these data. We attribute this to the fact that while we try to keep layers of the same size with the two models, Transformers lead to larger models, roughly 32M vs. 24M parameters, which can't probably be trained effectively on our small amount of data. Additionally, the character-level data format used as input to our models (please refer to \newcite{gugliotta-etal-2020-multi}) creates structured information on which LSTMs are notoriously more effective due to their computational power~\cite{weiss-etal-2018-practical,hahn-2020-theoretical}.

Finally, we compare different final steps experiments in table~\ref{tab:TarcLemmatization}. The first two, right below \emph{Step6}, are performed keeping the lemma as the last information level like in the annotation phase. In the first (\emph{global-split}), randomization is performed at sentence level at whole corpus level. In the second (genre-split), randomization is performed like in the previous section.
The latter leads to improvements on all levels except for POS, we keep thus this randomization strategy for the following experiments.
In the bottom block of table~\ref{tab:TarcLemmatization} we report results from models using the lemmas as the second level of information (note the new header for the bottom block). As we already mentioned in previous sections, the lemma should allow disambiguating the Arabic script and POS prediction, thus it should be put before them in the decoder's cascade.
As we can see, results are further improved on POS tagging and lemmatization (respectively 81.60 and 81.81 vs. 80.95 and 81.40).
Overall results are slightly worse than the corresponding ones in table~\ref{tab:MADATarc_AccIterativeResults}. While we find this surprising, we think this can be due to the fact that, while adding lemmas should improve overall results, it requires an additional decoder and thus additional parameters to train, in addition to the negative effect mistakes on the lemma level can have on transcription in CODA and tokenization levels.

\section{Conclusion}\label{Conclusion}

In this article we presented the outcomes of a three-year project that resulted in the creation of two tools to support research on Tunisian Arabic: a corpus of Tunisian Arabic encoded in Arabizi, namely a writing system for informal digital texts, and a neural network architecture created to annotate the corpus at various levels of linguistic information. We discussed the choices we made in terms of computational and linguistic methodology and the strategies we adopted to improve our results. We also reported some of the experiments that helped us to decide our path, by optimizing the available resources. Finally, we explained the reasons why we believe in the potential of these tools for both linguistic and computational research.

\section{Acknowledgements}

This work was supported by the CREMA project (\emph{Coreference REsolution into MAchine translation}) from the French National Research Agency (ANR), contract number ANR-21-CE23-0021-01.



\section{Bibliographical References}\label{reference}


\bibliographystyle{lrec2022-bib}
\bibliography{bibliography}


\end{document}